\documentclass[journal]{IEEEtran}
\usepackage{amssymb}
\usepackage{graphicx, amssymb, amsmath, booktabs, url}
\graphicspath{{Figure/}}

\begin{document}
%
\title{3D Randomized Connection Network with Graph-based Label Inference}
%
%
%

\author{Siqi Bao*, Pei Wang, Tony C. W. Mok and Albert C. S. Chung
\thanks{This work was supported in part by the Hong Kong Research Grants Council under Grant 16203115.}
\thanks{*S. Bao, P. Wang, T. Mok and A. Chung are with the Lo Kwee-Seong Medical Image Analysis Laboratory, Department of Computer Science and Engineering, The Hong Kong University of Science and Technology, Clear Water Bay, Hong Kong (e-mail: sbao@cse.ust.hk).}}

\maketitle

\begin{abstract}
In this paper, a novel 3D deep learning network is proposed for brain MR image segmentation with randomized connection, which can decrease the dependency between layers and increase the network capacity. The convolutional LSTM and 3D convolution are employed as network units to capture the long-term and short-term 3D properties respectively. To assemble these two kinds of spatial-temporal information and refine the deep learning outcomes, we further introduce an efficient graph-based node selection and label inference method. Experiments have been carried out on two publicly available databases and results demonstrate that the proposed method can obtain competitive performances as compared with other state-of-the-art methods.
\end{abstract}


%
\IEEEpeerreviewmaketitle

\section{Introduction}
The analysis of sub-cortical structures and pathological regions in brain Magnetic Resonance (MR) images is crucial in clinical diagnosis, treatment plan and post-operation assessment. Taking the Hippocampus for an example, the segmentation of this sub-cortical structure in brain MR images has been employed to predict the progression of Alzheimer's disease (AD). AD is the 6th leading cause of deaths in the United States, and it is estimated that there are approximately 5.5 million Americans living with AD in 2017 \cite{usad}. Besides the image segmentation of brain anatomical structures, segmenting some pathological regions, such as ischemic stroke lesion, is also invaluable in clinical decisions. Stroke is the 5th leading cause of deaths in the United States and kills more than 130,000 Americans each year \cite{usstroke}. The labeling difficulties stem from the irregularity of stoke lesion shape and unpredictability of its location, which makes it difficult to model its shape and acquire prior knowledge about its location. 

Recently deep learning techniques, such as Convolutional Neural Networks (CNN), have brought significant improvements in image labeling. The techniques evolve from image classification to semantic segmentation. For general image classification, it makes an inference about the image category of the input image based on achieved abstraction, i.e., assigning one specific label to the whole image. A large number of research works have been done to improve the classification accuracy \cite{krizhevsky2012imagenet, christian2014going, simonyan2014very} and some algorithms can even approach or outperform human beings \cite{lu2014surpassing, he2015delving}. As for image (semantic) segmentation, a correct label has to be assigned to each pixel based on the learned features. The elegant classification networks can help with the pixel label estimation by sliding the input patch across the image. It is a conventional and accurate way to predict the label for the center pixel based on the content abstractions from the patch \cite{farabet2013learning, pinheiro2013recurrent}. However, these patch-wise methods suffer from the expensive computation burden due to the dense prediction. To deal with these problems, the trick of shifting input and interlacing output was introduced in OverFeat \cite{sermanet2013overfeat}, which applies convolution kernels directly on the whole image rather than fix-sized patches. Some other image-wise methods for image segmentation have recently been proposed based on fully convolutional network (FCN) \cite{long2015fully, chen2014semantic}, by transforming the fully-connected layers in pre-trained classification network into convolutional layers.

Despite the progress of CNN in general image analysis, it is still challenging to apply these methods directly into brain MR image analysis, as these medical images are usually 3D volumes with poor contrast condition. To utilize CNN on 3D image analysis, the conventional way applies the 2D CNN network on each image slice (axial plane), and then concatenates the results along third image direction. Directly applying 2D convolution on 3D volumes will make the temporal information collapsed during the convolution process. To learn spatio-temporal features, 3D convolution is recently introduced in video analysis tasks \cite{ji20133d,tran2015learning}. Given the expensive computation cost, the size of convolution kernels is usually set to a small number in practice, which can only capture short-term dependencies.

For image segmentation with CNN, the classic architecture is fully convolutional network (FCN) \cite{long2015fully}. Due to the large receptive fields and pooling layers, FCN tends to produce segmentations that are poorly localized around object boundaries. Therefore, the deep learning outcomes are usually combined with probabilistic graphical models to further refine the segmentation results. Fully connected CRF \cite{koltun2011efficient} is one commonly used graphic model during the FCN post-processing \cite{chen2014semantic}, where each image pixel is treated as one graph node and densely connected to the rest graph nodes. Rather than utilizing the color-based affinity like fully connected CRF, boundary neural fields (BNF) \cite{bertasius2016semantic} first predicts object boundaries with FCN feature maps and then encodes the boundary information in the pair-wise potential to enhance the semantic segmentation quality. However, given the massive pixel amount and poor contrast condition in brain MR images, it is different to apply these methods directly to 3D brain image segmentation. 

To address the above challenges, in this paper, we extract long-term dependencies in spatial-temporal information with convolutional LSTM \cite{xingjian2015convolutional,patraucean2015spatio}. One novel randomized connection network is designed, which is a dynamic directed acyclic graph with symmetric architecture. Through the randomized connection, the deep network behaves like ensembles of multiple networks, which reduces the dependency between layers and increases the network capacity. To obtain the comprehensive properties for 3D brain image, both convolutional LSTM and 3D convolution are employed as the network units to capture long-term and short-term spatial-temporal information independently. Their results are assembled and refined together with the proposed graph-based node selection and label inference. Experiments have been carried out on the publicly available databases and our method can obtain quality segmentation results. 

Note that the preliminary version of this work has be presented in the 3rd Workshop on Deep Learning in Medical Image Analysis, in conjunction with MICCAI 2017. In this paper, 1) we extend our previous work by introducing the design of randomized connection and network units in detail; 2) additional mathematical equations, solutions together with illustrative examples are given in this work; 3) intensive experiments have been carried out to evaluate each component of our proposed method and comprehensive evaluations have been done with the state-of-the-art methods.

\section{Methodology}
In this section, we first introduce two kinds of network units: 3D convolution and convolutional LSTM, to capture short-term and long-term spatial-temporal information respectively. Then one novel symmetric network with randomized connection is presented as the architecture design. Graph-based node selection and label inference are further proposed to refine the labeling results efficiently.

\subsection{3D Convolution}
\begin{figure}
	\centering
	\includegraphics[width=\linewidth]{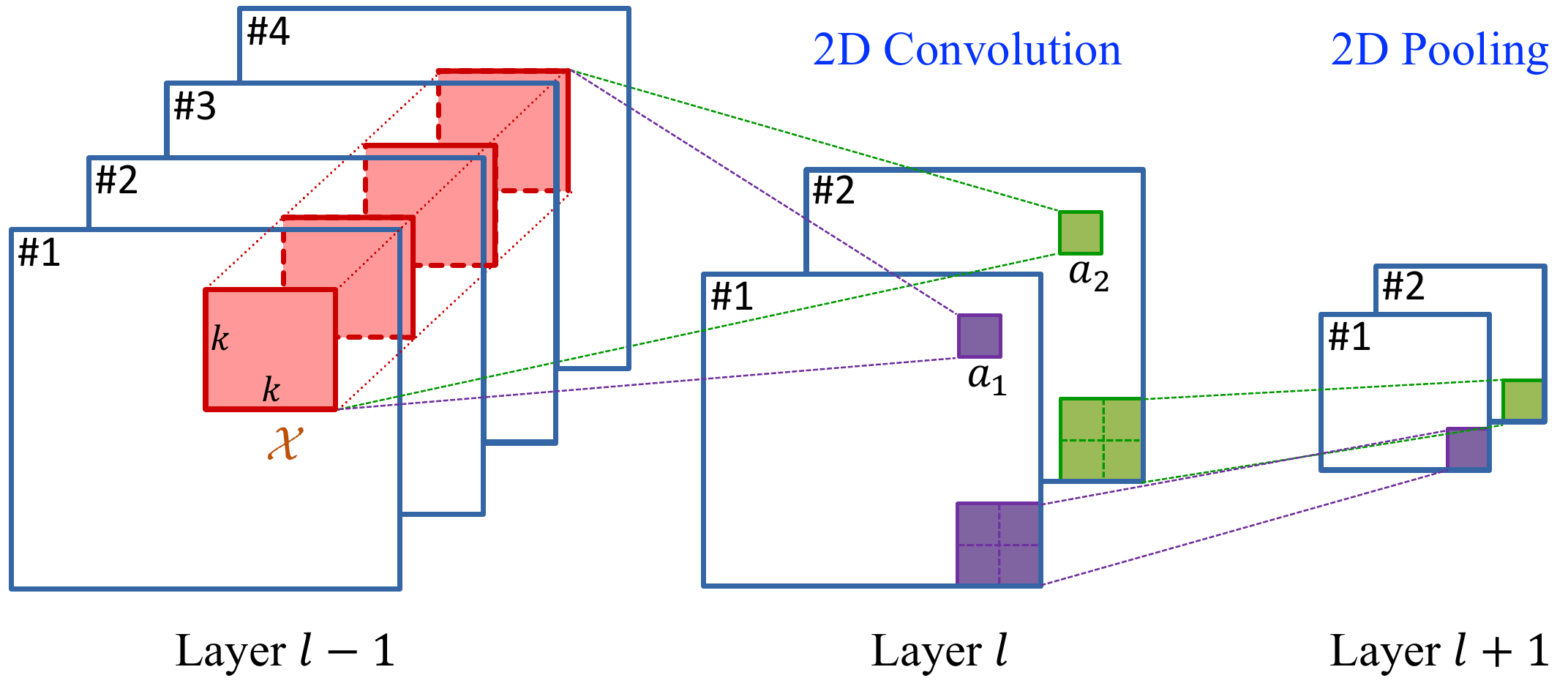}
	\caption{Illustration of 2D convolution with a kernel size of $k\times k$ and 2D pooling with a kernel size of $2\times 2$.}
	\label{fig1}
\end{figure}

Convolutional Neural Network (CNN) is a widely used deep learning technique in computer vision tasks, such as image classification, object detection and semantic segmentation. There are two basic components in CNN: convolution and pooling layer, as shown in Fig. \ref{fig1}. To compute the pixel values in one layer, those pixels within the corresponding local region from its last layer (namely receptive field) are employed as input. For example, the convolutional response $a$ in layer $l$ can be estimated as follows:
\begin{equation}\label{conv1}
a=f(W*\mathcal{X}+b),
\end{equation}
where $\mathcal{X}$ is the input from the receptive field (Red region in layer $l-1$), $W$ is the weight matrix and $b$ is the bias associated with the convolutional kernel. As for the non-linear activation $f(\cdot)$, it can be a traditional sigmoid or hyperbolic tangent function, or Rectified Linear Unit (ReLU) \cite{nair2010rectified}. As displayed in Fig. \ref{fig1}, one pair of $W$ and $b$ is corresponding to one feature map, and each feature map only has one single image. The receptive field size for convolution layer is $m\times k\times k$, where $m$ is the number of feature maps in the previous layer and $k$ represents the 2D convolutional kernel size. Since the convolutional kernels operate on a local neighborhood rather a single pixel, the spatial information can be captured and encoded in CNN. To obtain a more abstract feature representation, pooling layer is usually placed after convolution layer and the pooling strategy can be maximum or average pooling. From Fig. \ref{fig1}, it can be noticed that the pooling operation can only shrink the size of feature maps, while leave their amounts unchanged. 

\begin{figure}
	\centering
	\includegraphics[width=0.8\linewidth]{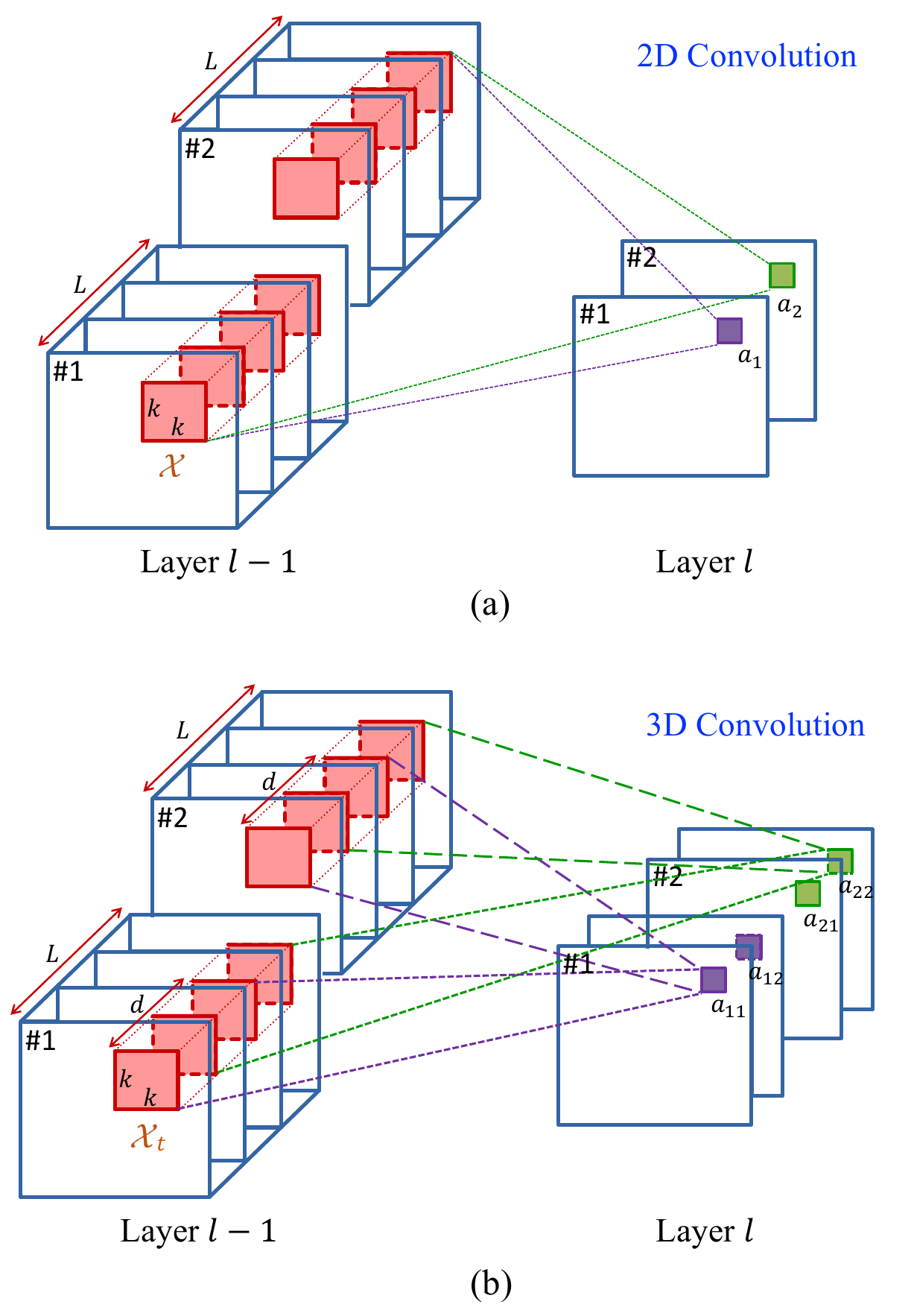}
	\caption{Distinction between 2D and 3D convolution with multiple volumes as input. In 2D convolution, multiple volumes result in one single image in each feature map, with the temporal information lost. In 3D convolution, multiple volumes result in multiple images in each feature map, with the temporal property reserved.}
	\label{fig2}
\end{figure}
To utilize CNN on 3D image analysis or video processing, the conventional way is first to apply the 2D CNN network on each image slice or frame, and then to concatenate the results along the third image direction or the time axis. Directly applying 2D convolution on 3D volumes will lead to the collapse of temporal information, since all frames (images) in the previous layer will result in one image. To learn spatial-temporal features, 3D convolution is recently introduced in video analysis tasks \cite{ji20133d, tran2015learning}. The distinction between 2D and 3D convolution is illustrated in Fig. \ref{fig2}. With 2D convolution, the size of receptive field $\mathcal{X}$ is $m\times L\times k\times k$, where $m$ is the number of feature maps in the previous layer and $L$ is the number of images in each feature map. Using 3D convolution, the size of receptive field $\mathcal{X}_t$ becomes $m\times d\times k\times k$, where $d$ is along the third image direction (time axis) and $d<L$. The 3D convolutional response $a_t$ in layer $l$ can be estimated in the following way:
\begin{equation}
a_t=\max(W*\mathcal{X}_t+b, 0).
\end{equation}
In this paper, we employ the ReLU $\max(\cdot, 0)$ as the non-linear activation $f(\cdot)$ in 3D convolution.

For both 2D and 3D convolutions, one pair of $W$ and $b$ is still corresponding to one feature map. While after 2D convolution, each feature map has only one single image (as shown in Fig. \ref{fig2}(a), Layer $l$), which leads to the loss of temporal information. For the feature map after 3D convolution, it still has multiple images and can keep tracking of temporary property. However, due to the expensive computation, the value of $d$ is usually assigned with a small number in practice ($d\times k\times k$ is often set to $3\times 3\times 3$ in 3D convolution, $d\ll L$), which is suitable to capture short-term dependencies. 

\subsection{Convolutional LSTM}
\begin{figure}
	\centering
	\includegraphics[width=0.85\linewidth]{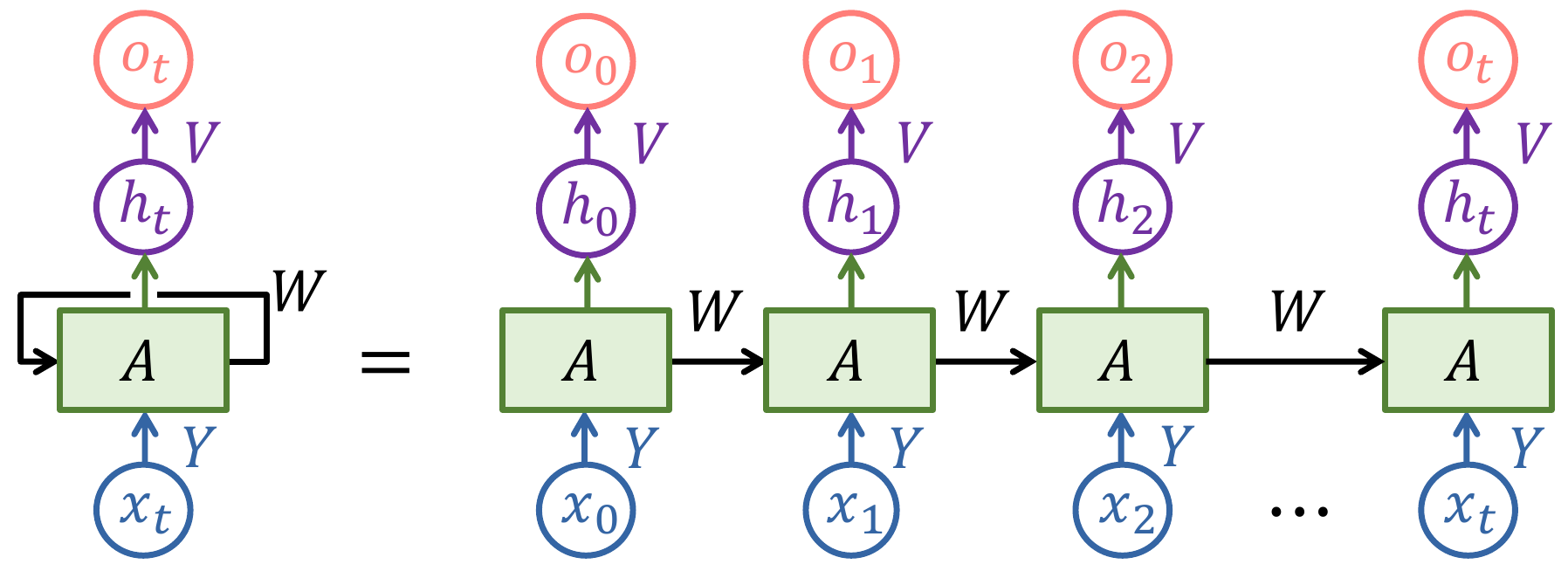}
	\caption{Left: Illustration of RNN with input $\{x_t\}_{t=1}^T$, hidden states $\{h_t\}_{t=1}^T$ and output $\{o_t\}_{t=1}^T$. $A$ is a neural network block. $Y$, $W$, and $V$ denote the input-to-hidden, hidden-to-hidden, and hidden-to-output weight matrices. Right: unrolled and equivalent RNN.}
	\label{fig3}
\end{figure}

Recurrent Neural Network (RNN) is another popular approach to collect temporal information, which is widely used in speech recognition and natural language processing. As displayed in Fig. \ref{fig3}, there is a loop inside RNN, which makes it inherently suitable for sequential modeling. To estimate the current hidden states $h_t$, it depends on both the current input $x_t$ and the previous hidden states $h_{t-1}$:
\begin{equation}\label{rnn1}
h_t=f(Wh_{t-1}+Yx_t+b).
\end{equation}
If the output $o_t$ is required, it can be calculated as follows:
\begin{equation}\label{rnn2}
o_t=Vh_t+c.
\end{equation}
$Y$, $W$, and $V$ denote the input-to-hidden, hidden-to-hidden, and hidden-to-output weight matrices, and $b$ and $c$ are the corresponding biases. Analogous to CNN, $f(\cdot)$ is the non-linear activation function. Although the previous hidden state information is encoded in RNN, it is incapable of modeling long-term dependencies in long sequences, since the signal decreases exponentially over time steps \cite{bengio1994learning}. Moreover, RNN suffers from the problem of gradient vanishing or exploding, which makes the optimization difficult.

\begin{figure}
	\centering
	\includegraphics[width=\linewidth]{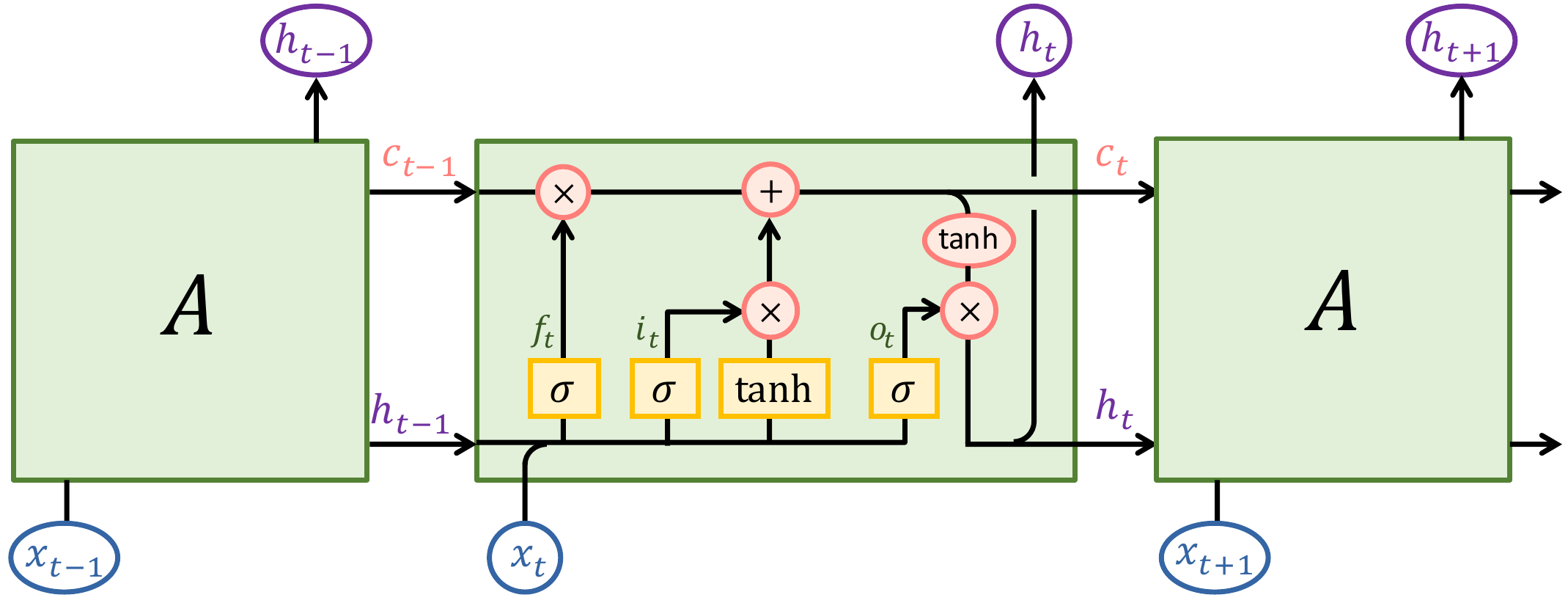}
	\caption{Illustration of LSTM: memory cell state $c_t$, forget gate $f_t$, input gate $i_t$ and output gate $o_t$.}
	\label{fig4}
\end{figure}

To deal with the above problem, Long-Short Term Memory (LSTM) \cite{hochreiter1997long} is proposed with a more complex neural network block to control information flow in a special way, as demonstrated in Fig. \ref{fig4}. The key component in LSTM is the memory cell state $c_t$, which carries information through the entire chain with some minor linear operations. This memory cell can be accessed and updated by three gates: forget gate $f_t$, input gate $i_t$ and output gate $o_t$. The forget gate $f_t$ decides how much information to be thrown away from the past cell state $c_{t-1}$ and the input gate $i_t$ determines the information to be accumulated into the latest cell state $c_t$. As for the output gate $o_t$, it controls the information propagation from the memory cell to the hidden state $h_t$.  Their detailed formulations are given as follows:
\begin{equation}\label{fc-lstm}
\begin{split}
& i_t=\sigma(W_{xi}x_t+W_{hi}h_{t-1}+b_i),\\
& f_t=\sigma(W_{xf}x_t+W_{hf}h_{t-1}+b_f),\\
& c_t=f_t\circ c_{t-1}+i_t\circ \tanh(W_{xc}x_t+W_{hc}h_{t-1}+b_c),\\
& o_t=\sigma(W_{xo}x_t+W_{ho}h_{t-1}+b_o),\\
& h_t=o_t\circ \tanh(c_t),
\end{split}
\end{equation}
where $\sigma(\cdot)$ and $\tanh(\cdot)$ refer to the sigmoid and hyperbolic tangent functions respectively. The symbol $\circ$ stands for Hadamard product, $W_{x\cdot}$, $W_{h\cdot}$ and $b_{\cdot}$ denote the input weights, recurrent weights and biases respectively. Because of the memory cell and the gating mechanism, during back-propagation, the error can be trapped inside the memory cell (also referred as constant error carousels \cite{hochreiter1997long}) through many time steps and the gradient can be prevented from vanishing or exploding quickly. 

In classic LSTM, fully-connected transformations are employed during the input-to-state and state-to-state transitions. As such, the spatial property is ignored. To gather the spatial-temporal information, convolutional LSTM (ConvLSTM) is recently proposed \cite{xingjian2015convolutional,patraucean2015spatio} to replace the fully-connected transformation with the local convolution operation. 

\begin{figure}
	\centering
	\includegraphics[width=\linewidth]{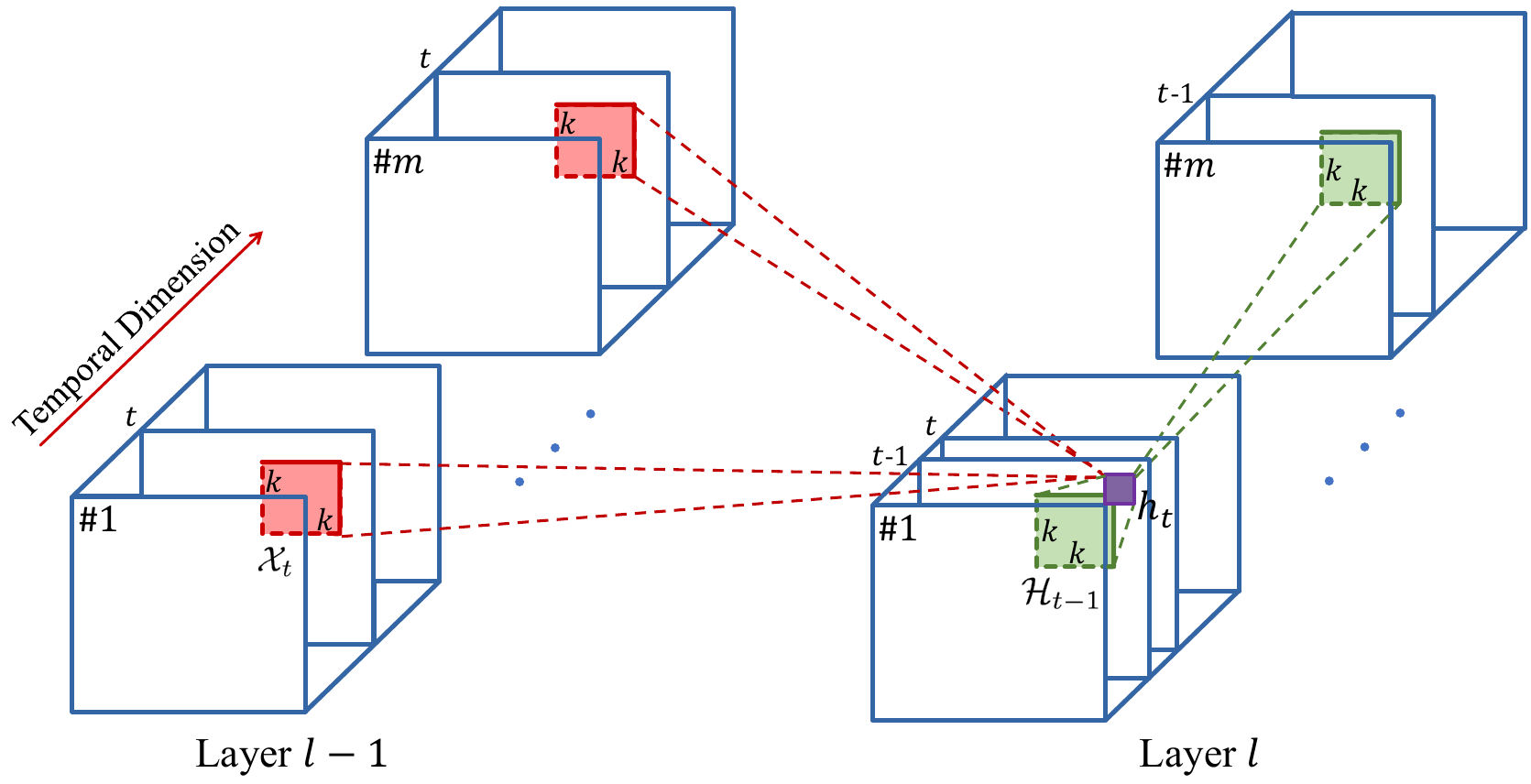}
	\caption{Illustration of convolutional LSTM. To compute the ConvLSTM response $h_t$ in layer $l$ (Purple pixel), both those pixels within the corresponding local region from its last layer (at the same time stamp, Red regions) and those from the current layer (at the previous time stamp, Green regions) are employed as input.}
	\label{fig7}
\end{figure}
In this paper, we utilize ConvLSTM to collect the long-term dependencies for 3D images, where the third image axis is treated as temporal dimension. The ConvLSTM for 3D image processing is illustrated in Fig. \ref{fig7}. To compute the pixel values in one layer, both those pixels within the corresponding local region from its last layer (at the same time stamp) and those from the current layer (at the previous time stamp) are employed as input. For example, the ConvLSTM response $h_t$ in layer $l$ (Purple pixel) can be estimated as follows: 
\begin{equation}\label{conv-lstm}
\begin{split}
& i_t=\sigma(W_{xi}*\mathcal{X}_t+W_{hi}*\mathcal{H}_{t-1}+b_i),\\
& f_t=\sigma(W_{xf}*\mathcal{X}_t+W_{hf}*\mathcal{H}_{t-1}+b_f),\\
& c_t=f_t\circ c_{t-1}+i_t\circ \tanh(W_{xc}*\mathcal{X}_t+W_{hc}*\mathcal{H}_{t-1}+b_c),\\
& o_t=\sigma(W_{xo}*\mathcal{X}_t+W_{ho}*\mathcal{H}_{t-1}+b_o),\\
& h_t=o_t\circ \tanh(c_t),
\raisetag{1\baselineskip}
\end{split}
\end{equation}
where $*$ denotes the convolution operation, the symbol $\circ$ stands for Hadamard product, $\sigma(\cdot)$ and $\tanh(\cdot)$ refer to the sigmoid and hyperbolic tangent functions respectively. As shown in Fig. \ref{fig7}, $\mathcal{X}_t$ is the input from last layer at the same time stamp (Red regions) and $\mathcal{H}_{t-1}$ is the input from current layer at the previous time stamp (Green regions). $W_{x\cdot}$ and $W_{h\cdot}$ denote the input-to-hidden and hidden-to-hidden weight matrices, with $b_{\cdot}$ as the corresponding biases. Distinct with the weight matrices in classical LSTM, the input $W_{x\cdot}$ and recurrent weights $W_{h\cdot}$ in ConvLSTM are all 4D tensors, with a size of $n\times m \times k \times k$ and $n\times n\times k\times k$ respectively, where $n$ is the predefined number of convolution kernels (feature maps in the current layer), $m$ is the number of feature maps in the previous layer and $k$ is the convolutional kernel size. ConvLSTM can be regarded as a generalized version of classic LSTM, with the last two tensor dimensions equal to $1$. 


\subsection{Randomized Connection Network}
With 3D convolution and ConvLSTM settled as network units to capture comprehensive spatial-temporal information, the next consideration is the design of the whole network architecture. Fully convolutional network (FCN) \cite{long2015fully} is a classic deep learning network for image segmentation, by transforming the fully-connected layers in pre-trained classification network into convolutional layers. To extract abstract features, poolings operations are indispensable in FCN, which leads to the significant size difference between estimated probability map and the original input image. It is necessary to employ extra up-sampling or interpolation steps to make up the size difference, while the segmentation quality through one direct up-sampling can be unacceptably rough. To address this problem, the network architecture of FCN turns from a line topology into a directed acyclic graph (DAG), by adding links to append lower layers with higher resolution into the final prediction layer. U-Net \cite{ronneberger2015u}, is another DAG with symmetric contracting and expanding architecture, which has gained great success in biomedical image segmentation. 3D U-Net \cite{cciccek20163d} is recently introduced for volumetric segmentation by replacing 2D convolution with 3D convolution. 

\begin{figure}
	\centering
	\includegraphics[width=\linewidth]{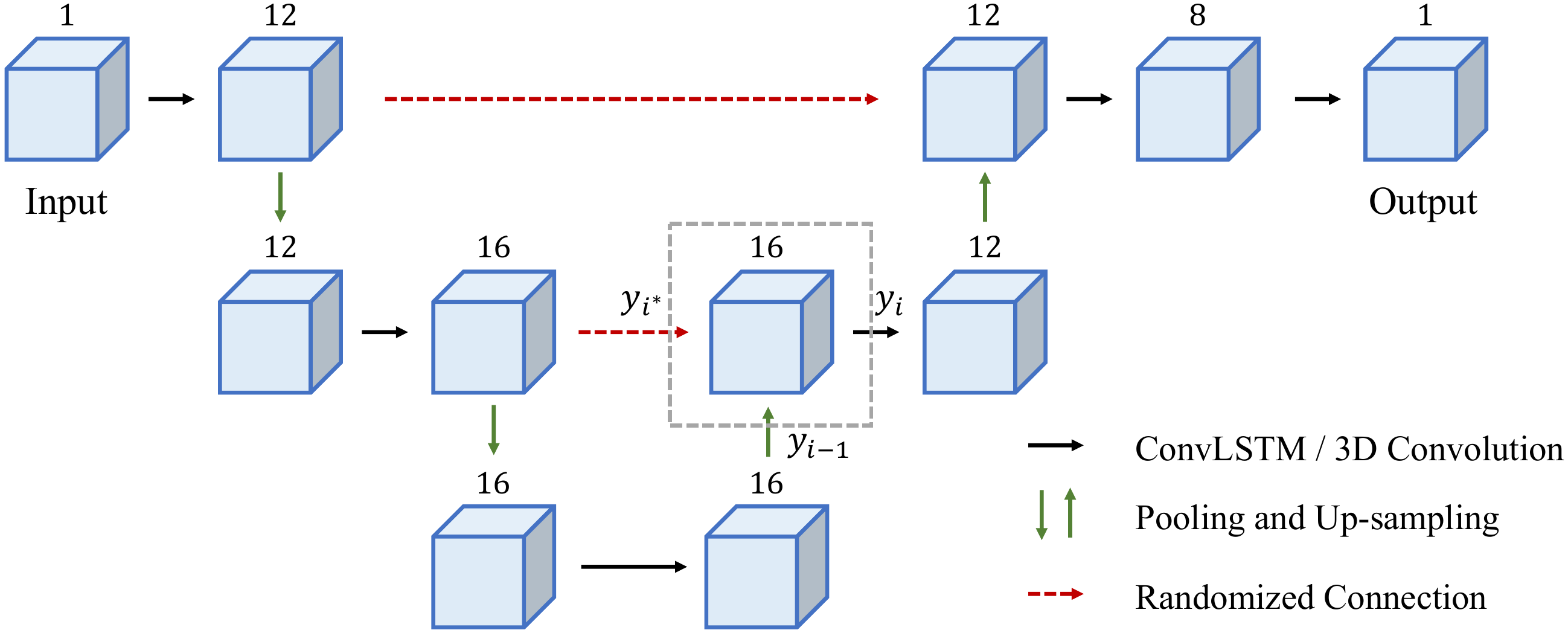}
	\caption{Illustration of 3D Randomized Connection Network. The numbers above cubes refer to the number of feature maps in that layer. Gray dashed square is for the concept explanation of randomized connection. }
	\label{fig8}
\end{figure}
Inspired by the improvements in biomedical image analysis using U-Net, in this paper, we also keep the symmetric contracting and expanding structure for 3D brain image segmentation, with detailed network shown in Fig. \ref{fig8}. The 3D convolution/ConvLSTM (Black arrow) is employed to capture the short-term or long-term spatial-temporal properties. The Green arrows refer to the pooling or upsampling operations. Distinct with U-Net where all connections are fixed (static DAG), in the proposed method, the connection between contracting and expansive paths (Red arrow) is randomly established during training (dynamic DAG). To further illustrate the concept, we use one layer as an example to analyze its input and output. For the $i$-th layer with randomized connection (Grey dashed square) along the expansive path, its output can be estimated as:
\begin{equation}
y_i = \mathcal{U}(y_{i-1}) + \mathcal{R}(y_{i^*}, \alpha),
\end{equation}
where $y_{i-1}$ is the input from the previous layer along the expansive path, $\mathcal{U}(\cdot)$ is the upsampling operation, and $y_{i^*}$ the input from corresponding layer along the contracting path. $\mathcal{R}(y_{i^*}, \alpha)$ is a randomized function whose result is $y_{i^*}$ with the probability $\alpha$, and $0$ with the probability $1-\alpha$. During training, the input $y_{i^*}$ will be added to $i$-th layer with the probability $\alpha$ in each iteration. 

It is worth noting that randomized connection is different from dropout, although both of them are trying to enforce regularization on the deep networks to decrease overfitting during training. Dropout intends to prevent the co-adaptation of neurons in neural networks, by randomly selecting a subset of units and setting their outputs to zero. While the proposed randomized connection intends to reduce the dependency between layers and to increase the model capacity. By randomly dropping the summation connection, the layers can be fully activated and forced to learn instead of relying on the previous ones.

Randomized connection achieves great robustness and efficiency because it reduces dependency between layers and increases the model capacity. By randomly dropping the summation connection, the layers can be fully activated and forced to learn instead of relying on the previous ones. As discussed in \cite{veit2016residual}, residual network with identity skip-connections behaves like ensembles of relatively shallow networks. In the proposed method, the summation connection is randomly established in every iteration, so a number of different models are assembled implicitly during training. If there are $N$ connections linking the two paths, then it will be $2^N$ models combined in the training process. In the proposed method, two randomized connection networks are trained independently, with ConvLSTM and 3D convolution as network unit to capture long-term and short-term spatial-temporal information respectively. 

\subsection{Graph-based Label Inference}\label{graph_inf}
Distinct with general images, which usually are 2D images and have relatively sharp object boundaries, the size of medical volumes is much larger and the boundary among tissues is quite blurry as a result of the poor contrast condition. Although fully connected CRF and BNF can boost the segmentation performance for general images, this kind of differences in image properties might lead to some problems if directly applying these methods on 3D medical image segmentation. Given that the typical size for brain MR images \cite{shattuck2008construction} can be $256\times 256\times 124$, with fully connected CRF, the amount of connecting edges for one node becomes $8,126,463$. On one hand, dense connections on a huge graph can suffer from heavy computation burden during optimization. On the other hand, due to the similar histogram profiles among different tissues in medical images, the dense connection can incur extra outliers or generate spatially disjoint predictions. For the boundary-based method BNF, its application gets hindered by the poor contrast condition in brain images. As such, it is necessary to design effective graph-based inference method for 3D brain image segmentation. 

The proposed graph-based label inference method involves two steps: node selection and label inference. For the sake of efficiency, it is better to prune the majority of pixels and to focus on those whose results need to be refined. The node selection and label inference are introduced based on the fundamental graph $G=(V, E)$, where node set $V$ includes all pixels in the 3D image and edge set $E$ corresponds to the image lattice connection. If $v_i$ and $v_j$ are adjacent in the 3D image, an edge $e_{ij}$ will be set up, with $w_{ij}$ as edge weight. As both long-term and short-term spatial-temporal information are desirable in the node selection process, the labeling results estimated by ConvLSTM and those by 3D convolution need to be employed collaboratively. Note that the examples and figures in this subsection are just for simplification to use two network results. In fact, the node selection and label inference are not limited to the number of networks. 

For each node $v_i\in V$, it can be represented as $v_i=\{v_i^1, v_i^2, \cdots,v_i^K\}=\{(p_i^1,1-p_i^1),(p_i^2,1-p_i^2),\cdots,(p_i^K,1-p_i^K)\}$, where $p_i^k$ refers to the probability estimated by the $k$-th deep learning network that $v_i$ belongs to foreground. During node selection, two criteria are taken into consideration: the \textbf{label confidence} of each node and the \textbf{label consistency} among the neighborhood. We want to filter out those nodes with high label confidence and consistency, so that we can focus on the rest nodes for further processing. In Fig. \ref{fig3}, two small image cubes are extracted from two result images for illustration. For the node $v_i$ in the $k$-th result image (Yellow node), its confidence is evaluated by the contrast between foreground and background probability, with the definition as follows:
\begin{equation}
C_f(v_i^k)=(1-p_i^k-p_i^k)^2=(1-2p_i^k)^2.
\end{equation}
As for the consistency, it is measured by the cosine similarity between neighboring nodes, defined as:
\begin{equation}
C_s(v_i^k,v_j^{\cdot})=\cos(v_i^k,v_j^{\cdot})=\frac{{v_i^k}^T v_j^{\cdot}}{\|v_i^k\| \|v_j^{\cdot}\|},
\end{equation}
where $v_j^{\cdot}\in \mathcal{N}(v_i^k)$, $\mathcal{N}(\cdot)$ includes the 6-nearest neighbors in $k$-th result image (Blue node) and the corresponding nodes $v_i^{\cdot}$ in the rest of the images (Yellow node). 

\begin{figure}[t]
	\centering
	\includegraphics[width=\linewidth]{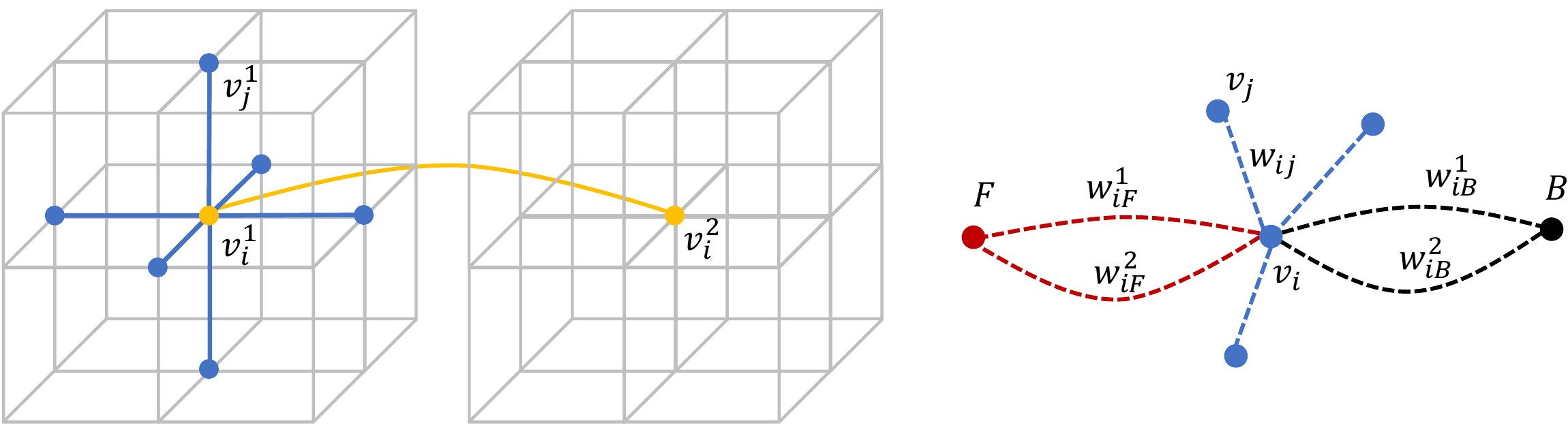}
	\caption{Graph-based Inference Illustration. \textbf{Left}: node selection (cubes extracted from 1st and 2nd result images). \textbf{Right}: graph-based label inference for candidate nodes.}
	\label{fig5}
\end{figure}
The two criteria are combined together for nodes selection and the detailed formulation is given as follows:
\begin{equation}\label{seed-sel}
\begin{split}
\max ~~&\sum_{v_i\in V^*}\sum_k (C_f(v_i^k)+\sum_{v_j^{\cdot}\in \mathcal{N}(v_i^k)}C_s(v_i^k,v_j^{\cdot})),\\
s.t. ~~&|V^*|\leq |V|\times \theta,
\end{split}
\end{equation}
where $\theta$ is the pre-defined threshold, indicating the percentage of nodes to be pruned, and $|V^*|$ is the set of confident nodes that can be pruned. The first unary term $C_f$ measures the label confidence and the second pair-wise term $C_s$ accesses the label consistency. Equation \eqref{seed-sel} can be solved efficiently by sorting the energy for each $v_i$ in descending order and then set the first $|V|\times \theta$ nodes as confident nodes. The rest of the nodes are treated as candidate nodes and need further label inference. 

\begin{table*}
	\caption{Experimental results on the LPBA40 database, measured with DC. Highest values are written in Red.}
	\centering
	\includegraphics[width=0.95\linewidth]{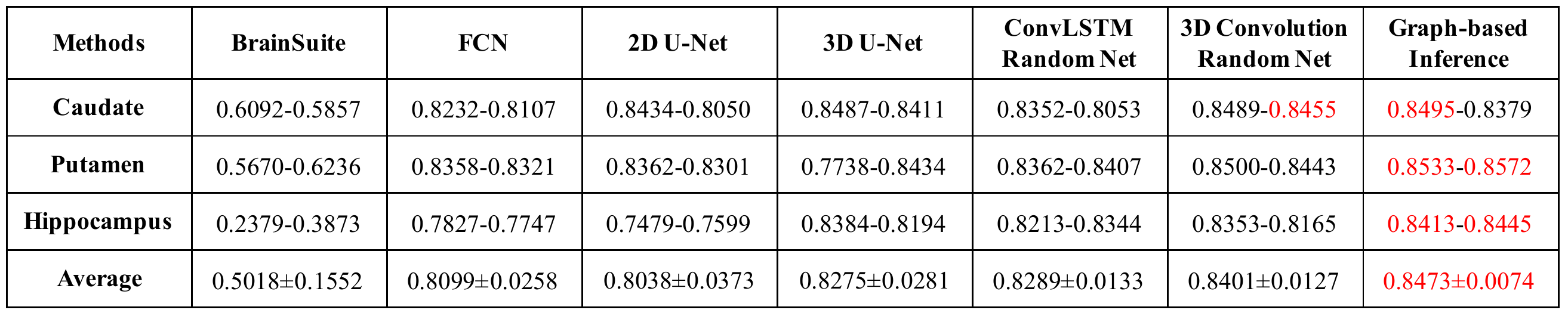}
	\label{result_lpba}
\end{table*}
The label inference is developed on a compact graph $G^\mathcal{C}=(V^\mathcal{C}, E^\mathcal{C})$, where $V^\mathcal{C}$ is candidate node set and $E^\mathcal{C}$ is the lattice edge connecting candidate nodes. The inference problem is formulated under the Random Walker framework \cite{grady2006random}, with detailed definition given as follows:
\begin{equation}\label{obj}
\begin{split}
\min_{x}& \sum _{v_i}\sum_k~[{w_{iF}^k}^2(x_i-1)^2+{w_{iB}^k}^2x_i^2]+\sum _{e_{ij}}~w_{ij}^2(x_i-x_j)^2,\\
s.t.~& x_F=1,~x_B=0,
\raisetag{1\baselineskip}
\end{split}
\end{equation}
where $x_i$ is the probability that node $v_i$ belongs to the foreground, $F$ and $B$ refers to the foreground and background seed respectively, as shown in Fig. \ref{fig3}. In the first unary term, $w_{iF}^k$ and $w_{iB}^k$ are the priors from deep learning network, which are assigned with $p_i^k$ and $1-p_i^k$. In the second pairwise term, $w_{ij}$ is the edge weight for lattice connection (Blue dashed line), which is estimated by conventional Gaussian function: 
\begin{equation}
w_{ij}=\exp(-\beta (I(v_i)-I(v_j))^2),
\end{equation}
where $I(\cdot)$ is the intensity value and $\beta$ is a tuning parameter. By minimizing Equation \eqref{obj}, the probability $x_i$ for each candidate node can be obtained and the label can be then updated correspondingly: $L(v_i) =1$ if $x_i\geq 0.5$ and $L(v_i) =0$ otherwise. 

\section{Experiments}
In this paper, experiments have been carried out on two publicly available brain MR image databases -- LPBA40 \cite{shattuck2008construction} and Ischemic Stroke Lesion Segmentation (ISLES) Challenge 2016 \cite{maier2017isles}. 

\subsection{Segmentation Results on LPBA40}
LPBA40 has 40 volumes with 56 structures delineated. The database was randomly divided into two equal sets for training and testing respectively. Data augmentation with elastic transformation was performed to increase the amount of training data by 20 times and the training process was set to 60 epochs, with a learning rate of $10^{-4}$. The rest of the parameter settings used in the experiments are listed as follows: the probability for randomized connection $\alpha=0.5$, the percentage to prune nodes $\theta=0.999$ and the tuning parameter in Gaussian function $\beta=100$. Standard dropout regularization has also been utilized for the proposed networks and all the compared methods in the experiments.

Recently several softwares have been available to provide the automatic segmentation function for brain MR images, such as BrainSuite \cite{shattuck2002brainsuite} or FreeSurfer \cite{fischl2012freesurfer}. During evaluation, we utilized BrainSuite, one of the available softwares, to segment images in the LPBA40 databases as a reference. BrainSuite first runs surface/volume registration based on the extra high-resolution ($0.5mm\times 0.5mm \times 0.8mm$) BCI-DNI\_brain atlas and then warps the label map from the atlas to the target image. In the experiments, FCN was employed as the baseline, where one patch-based classification network was first trained and then adapted to image-based segmentation network by transforming fully connected layer to convolutional layer. Besides the reference BrainSuite and the baseline FCN, we also compared with the state-of-the-art methods -- symmetric U-Net with fixed connection using 2D and 3D as network units. 

Dice Coefficient (DC) is utilized to measure the quality of segmentation results. The quantitative results on available sub-cortical structures are given in Table \ref{result_lpba}, with the highest values shown in Red. Each sub-cortical structure has two parts (located in the left and right hemisphere), and the results are provided for the left-right part respectively, separated by the hyphen. The intermediate results generated by randomized connection network using 3D Convolution and ConvLSTM are provided in this table. Although BrainSuite utilizes a high-resolution atlas, those deep learning based methods (FCN, U-Net and our method) which rely on the low-resolution atlases inside the database, obtain much better performances. 

In the experiments, the network architectures for 3D U-Net and 3D Convolution Random Net are kept the same, including the number of feature maps, kernel size and the employment of standard dropout, except the fixed connection and the proposed randomized connection. From the comparison of quantitative results between them, it shows that the proposed randomized connection can help improve the segmentation significantly by 1.26\% on the LPBA40 database. 

As compared with conventional FCN and U-Net, randomized connection networks can obtain better results. Through graph-based label inference, the long-term and short-term information can be assembled together to further improve the performance. Some visual results are shown in Fig. \ref{fig4}, with outliers circled in Red. The first column is the ground truth for reference, the second and third columns are results from ConvLSTM and 3D convolution randomized connection networks. As displayed in the fourth column, these outliers in randomized networks can be removed after graph-based label inference. 

\subsection{Segmentation Results on ISLES Challenge 2016}
In the ISLES Challenge 2016, it has two tasks: the segmentation of stroke lesion volumes and the regression of clinical mRM score. There are 30 cases in the training dataset and 19 cases in the testing dataset. In the experiments, we focus on the segmentation of ischemic stroke lesion regions. Data augmentation with elastic transformation was performed to increase the amount of training data by 20 times and the rest parameter settings were kept as the same with those in the LPBA40 database.  
\begin{table}
	\caption{Experimental results on the ISLES Challenge 2016, measured with DC. Our results are shown in Red.}
	\centering
	\includegraphics[width=0.95\linewidth]{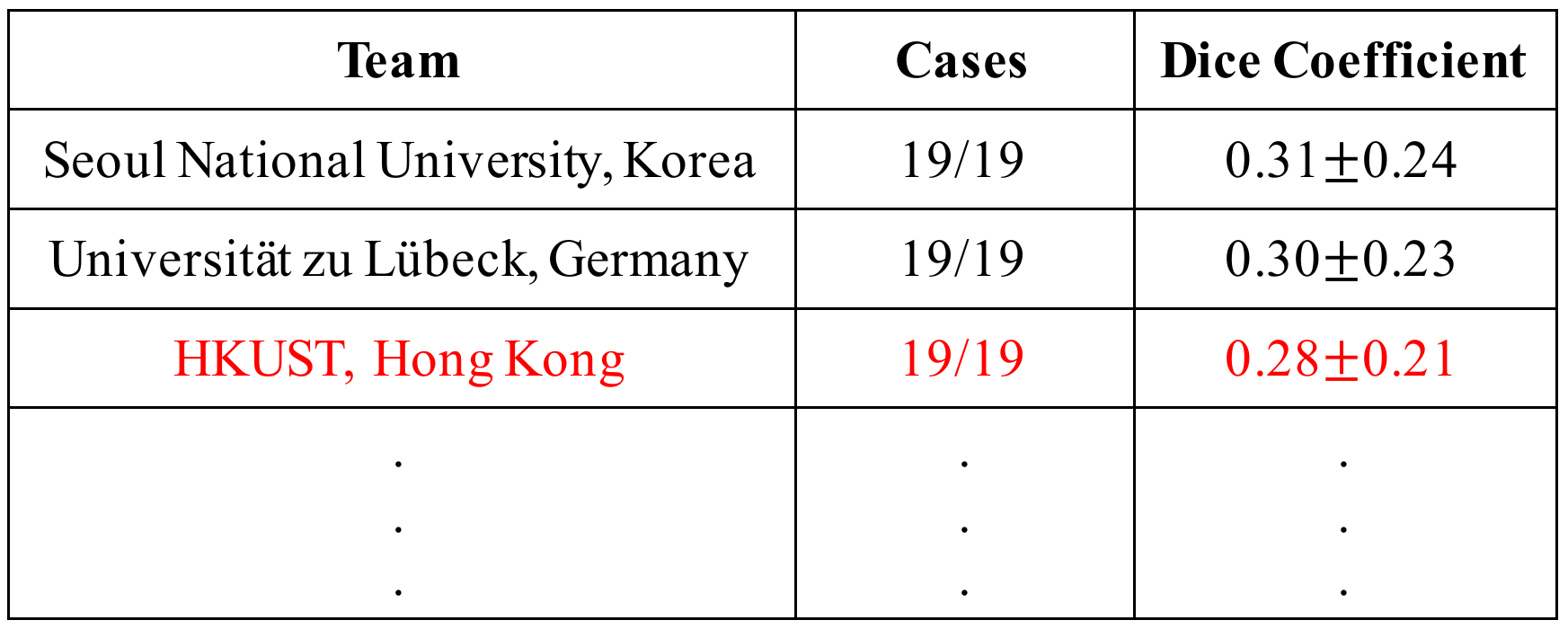}
	\label{result_isles}
\end{table}

\begin{figure*}
	\centering
	\includegraphics[width=0.98\linewidth]{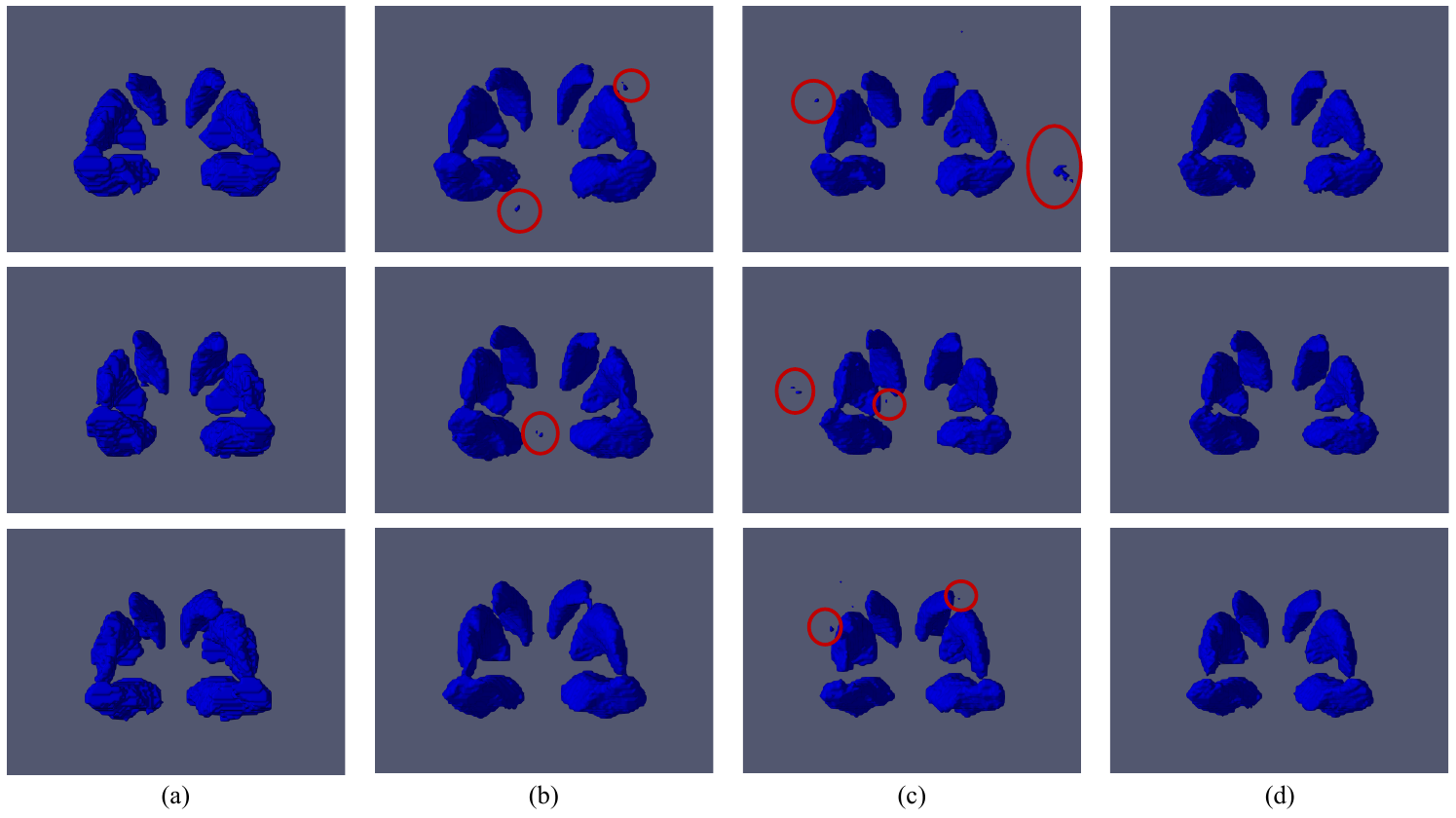}
	\caption{Some visual segmentation results of sub-cortical structures on the LPBA40 database. Each row represents the 3D labeled volumes for one subject. (a) Ground truth for reference; (b) Intermediate labeling results by ConvLSTM randomized connection network; (c) Intermediate labeling results by 3D convolution randomized connection network; (d) Final segmentation after graph-based label inference.}
	\label{fig6}
\end{figure*}
\begin{figure*}
	\centering
	\includegraphics[width=\linewidth]{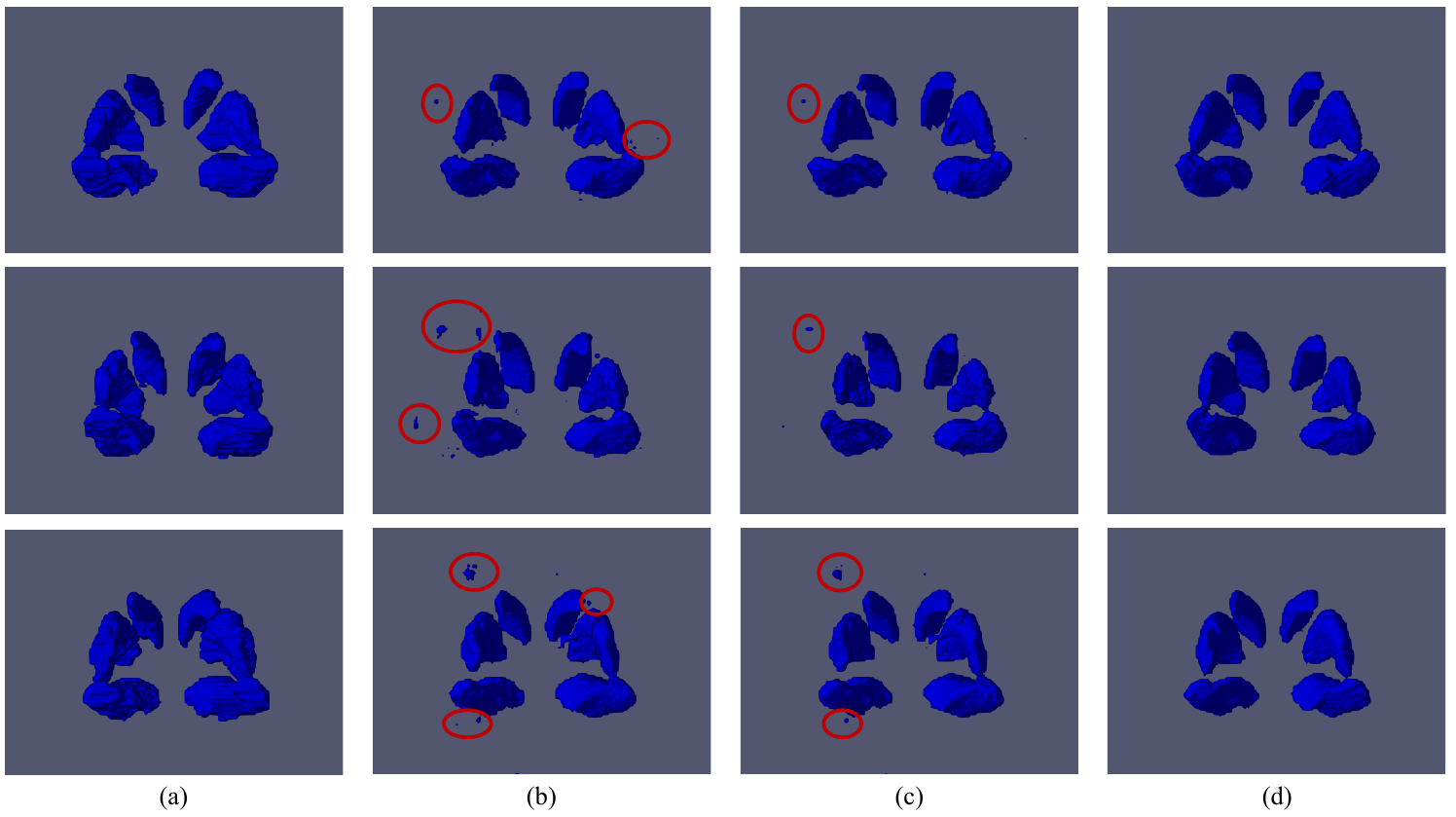}
	\caption{Some visual segmentation results by fully connected CRF on the LPBA40 database. Each row represents the 3D labeled volumes for one subject. (a) Ground truth for reference; (b) Labeling results by 3D U-Net; (c) Segmentation after fully connected CRF; (d) Segmentation by graph-based label inference for reference.}
	\label{fig11}
\end{figure*}
As compared with the labeling of sub-cortical structures, the segmentation of ischemic stroke lesion is more challenging, as the shape and position of pathological regions are not predictable. Many methods cannot successfully to label all the 19 cases in test dataset. In Table \ref{result_isles}, we list several teams which have finished the labeling of all the 19 cases, measured with DC and our team ranks the 3rd in this list. (Here we only consider the results which can finish all the labeling of 19 cases, while the list on the chanllendge website also includes other dice results which cannot finish the 19 cases.) As compared with the patch-wise segmentation method employed by the 1st team, the performance of our proposed method is competitive and our image-wise segmentation is more efficient as it can provide the labeling map for the 3D image directly.

\subsection{Further Discussion}
\begin{figure}
	\centering
	\includegraphics[width=\linewidth]{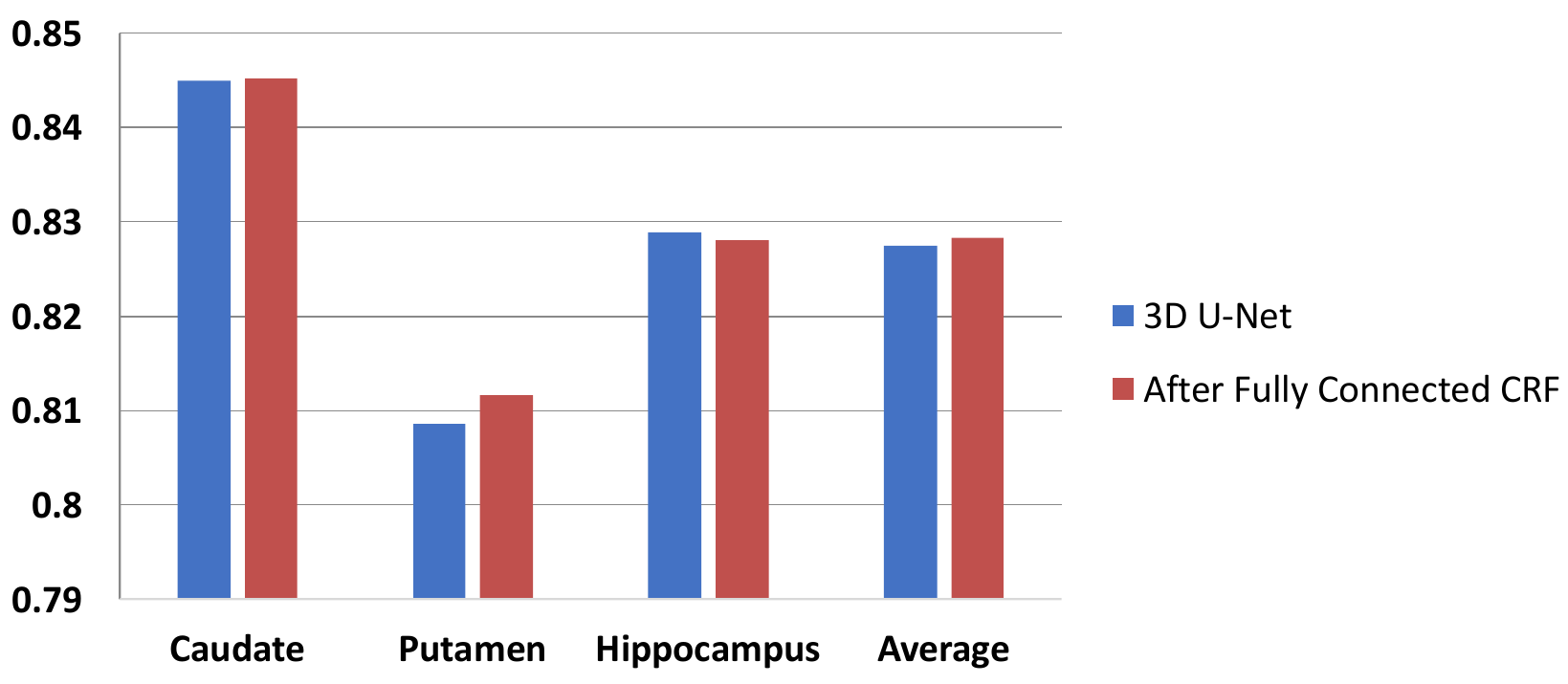}
	\caption{Segmentation quality improvements brought by fully connected CRF, measured with DC.}
	\label{fig10}
\end{figure}
For the evaluation of our proposed graph-based method, a small scale of experiments using fully connected CRF was performed to post-process deep learning outcomes by 3D U-Net on the LPBA40 database. We employed the publicly available implementation of fully connected CRF for processing 3D images \cite{koltun2011efficient, dense3dCrf}. Some visual results and the quality improvements brought by fully connected CRF measured with DC are given in Fig. \ref{fig11} and Fig. \ref{fig10}, respectively. As shown in Fig. \ref{fig11}, the first and fourth columns are the ground truth and our graph-based label inference result for reference. The second column displays the labeling results by 3D U-Net, and some outliers still exist even after the post-processing with fully connected CRF in the third column. This might be caused by the poor contrast condition and similar histograms among tissues in brain MR images (\textit{discussed in detail in Section \ref{graph_inf}}). From Fig. \ref{fig10}, we can observe that the improvements brought by fully connected CRF after 3D U-Net is limited, only 0.08\% measured by DC. As for the running time to label one sub-cortical structure in one target image, it takes around 1 minute using the fully connected CRF (on a 3.2GHz, Quad-Core CPU with 8GB RAM machine), as compared with 4 seconds using our graph-based label inference.

\begin{figure}
	\centering
	\includegraphics[width=\linewidth]{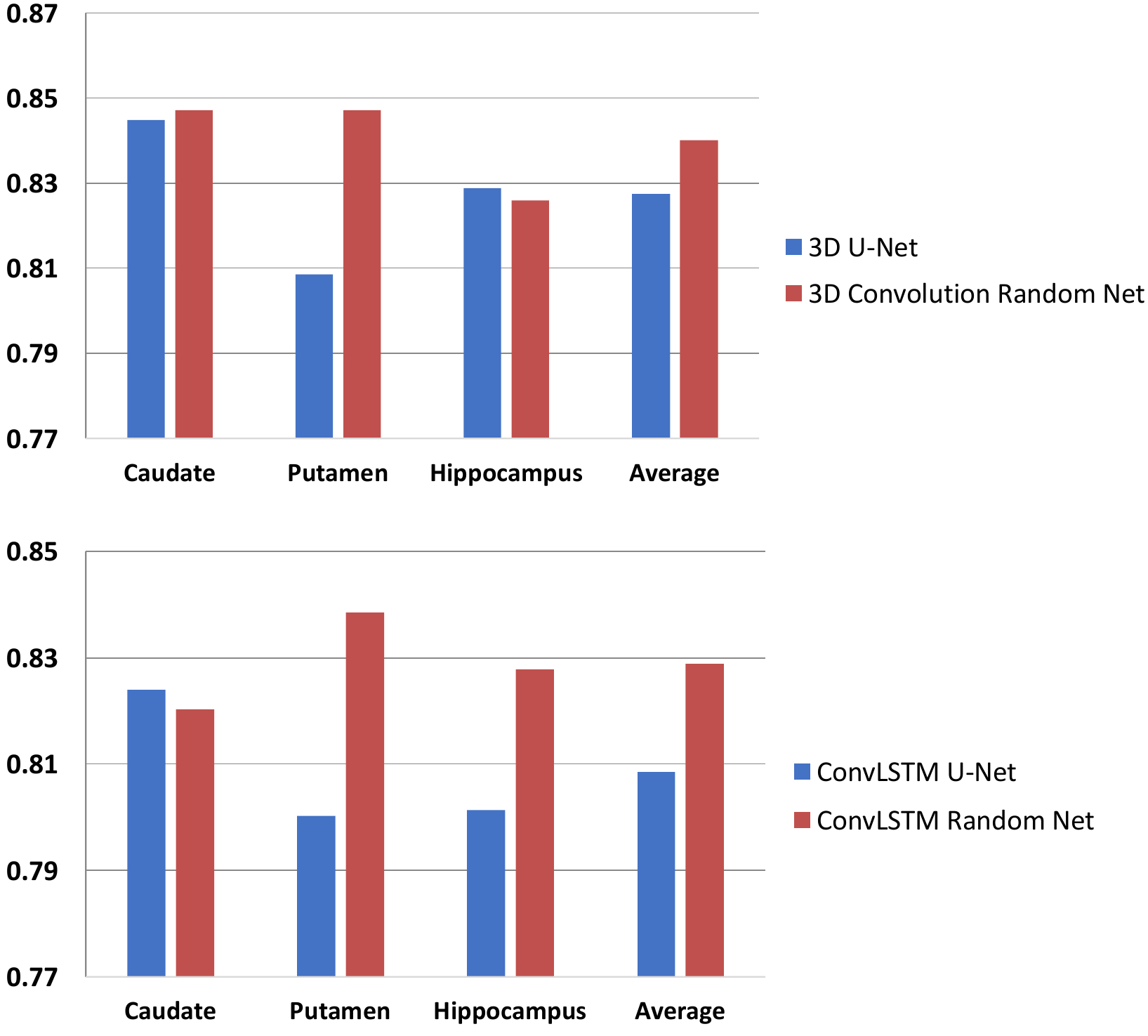}
	\caption{Effects of randomized connection. \textbf{Upper:} the comparison between 3D U-Net and 3D Convolution Random Net; \textbf{Bottom:} the comparison between ConvLSTM U-Net and ConvLSTM Random Net.}
	\label{fig9}
\end{figure}

To test the performance of our proposed randomized connection, the comparisons between randomized connection networks and corresponding symmetric U-Net with fixed connection are shown in Fig. \ref{fig9}. In the upper figure, the comparison between 3D U-Net and 3D Convolution Random Net indicates that the randomized connection can improve the labeling quality by 1.26\%. In the bottom figure, the comparison between ConvLSTM U-Net and ConvLSTM Random Net indicates that the randomized connection can improve the labeling quality significantly by 2.03\%.

\section{Conclusion}
In this paper, a novel deep network with randomized connection is proposed for 3D brain image segmentation, with ConvLSTM and 3D convolution network units to capture long-term and short-term spatial-temporal information respectively. The proposed randomized connection is able to enforce regularization on the deep networks to decrease overfitting during training, by controlling the connections among layers. To determine the label for each pixel efficiently, the graph-based node selection is introduced to prune the majority quality nodes and to focus on the nodes that really need further label inference. The long-term and short-term dependencies are encoded to the graph as priors and utilized collaboratively in the graph-based inference. Experiments carried out on the publicly available databases indicate that our method can obtain the competitive performance as compared with other state-of-the-art methods.


\ifCLASSOPTIONcaptionsoff
  \newpage
\fi

\cleardoublepage
\bibliographystyle{IEEEtran}
\bibliography{tip2017}

\end{document}